\documentclass[11pt]{article}
\usepackage{paclic35}
\usepackage{times}
\usepackage{latexsym}
\usepackage{amsmath}
\usepackage{multirow}
\usepackage{url}
\usepackage{float}
\usepackage{tablefootnote}
\usepackage{graphicx}
\usepackage{booktabs}
\usepackage{tabularx}

\title{Span Detection for Aspect-Based Sentiment Analysis in Vietnamese}

\author{Kim Thi-Thanh Nguyen$^{1,2,3}$, Sieu Khai Huynh$^{1,2,3}$, \textbf{Phuc Huynh Pham}$^{1,2,3}$, \textbf{Luong Luc Phan}$^{1,2,3}$,\\\textbf{Duc-Vu Nguyen}$^{1,2,4}$, \textbf{Kiet Van Nguyen}$^{1,2,4}$\\
$^{1}$University of Information Technology, Ho Chi Minh City, Vietnam\\
$^{2}$Vietnam National University, Ho Chi Minh City, Vietnam \\
$^{3}$\{18520963,18520348,18521260,18521073\}@gm.uit.edu.vn\\
$^{4}${\{vund, kietnv\}}@uit.edu.vn}

\date{}

\begin{document}
\maketitle
\begin{abstract}
Aspect-based sentiment analysis plays an essential role in natural language processing and artificial intelligence. Recently, researchers only focused on aspect detection and sentiment classification but ignoring the sub-task of detecting user opinion span, which has enormous potential in practical applications. In this paper, we present a new Vietnamese dataset (UIT-ViSD4SA) consisting of 35,396 human-annotated spans on 11,122 feedback comments for evaluating the span detection in aspect-based sentiment analysis. Besides, we also propose a novel system using Bidirectional Long Short-Term Memory (BiLSTM) with a Conditional Random Field (CRF) layer (BiLSTM-CRF) for the span detection task in Vietnamese aspect-based sentiment analysis. The best result is a 62.76\% F1$_{macro}$ for span detection using BiLSTM-CRF with embedding fusion of syllable embedding, character embedding, and contextual embedding from XLM-RoBERTa. In future work, span detection will be extended in many NLP tasks such as constructive detection, emotion recognition, complaint analysis, and opinion mining. Our dataset is freely available at \url{https://github.com/kimkim00/UIT-ViSD4SA} for research purposes.  

\end{abstract}

\section{Introduction}
\label{introduction}
Typically, before buying an item or deciding to use a service, people tend to seek advice from their predecessors who purchased the item or used the service. With the rapid development of the internet, more and more people find advice from websites, e-commerce sites, forums, or product review channels. Massive user reviews available on e-commerce platforms are becoming valuable resources for both customers and producers. For customers, this data source provides information about products and helpful advice to help them avoid buying products or signing up for services that are not suitable for their personal needs. On the other hand, user reviews are also valuable information for businesses, and if used correctly and effectively, this data can help businesses improve product quality, accurately identify the target customers for each segment.

Aspect-based sentiment analysis (ABSA) \cite{10.1145/1014052.1014073} on user feedback is a  challenging task that attracts interest from both academia and industries \cite{10.1145/1935826.1935932,kiritchenko-etal-2014-nrc,chen-etal-2017-recurrent-attention}. Given specific feedback about a product or service, the main task of ABSA is to detect what is being discussed, then give sentiment analysis to the explored aspect. The ABSA problem can be divided into three basic tasks as follows: aspect detection, opinion target expression (OTE), sentiment polarity. In this paper, we focus on detecting the opinions of users based on aspects and their sentiment, which we call span detection for ABSA. Specifically, when a review is given \textit{”Although staffs are nice, the phone is terrible!”}, the span detection for ABSA task aims to get two opinions \textit{"staffs are nice"} and \textit{"the phone is terrible"}, then classify these into right aspects also sentiment polarity. The task is described as follows: 
\begin{itemize}
    \item {\bf Input}: A customer feedback $C$ for a smartphone that consists of $n$ characters. 
    \item {\bf Output}: One or more spans of customer opinions are extracted directly from feedback $C$ for each aspect. Each span is extracted from position $i$ to position $j$ such that $0\leq i,j\leq n$ and $i \leq j$. 
\end{itemize}

\begin{figure}[ht]
    \centering
    \includegraphics[scale=0.39]{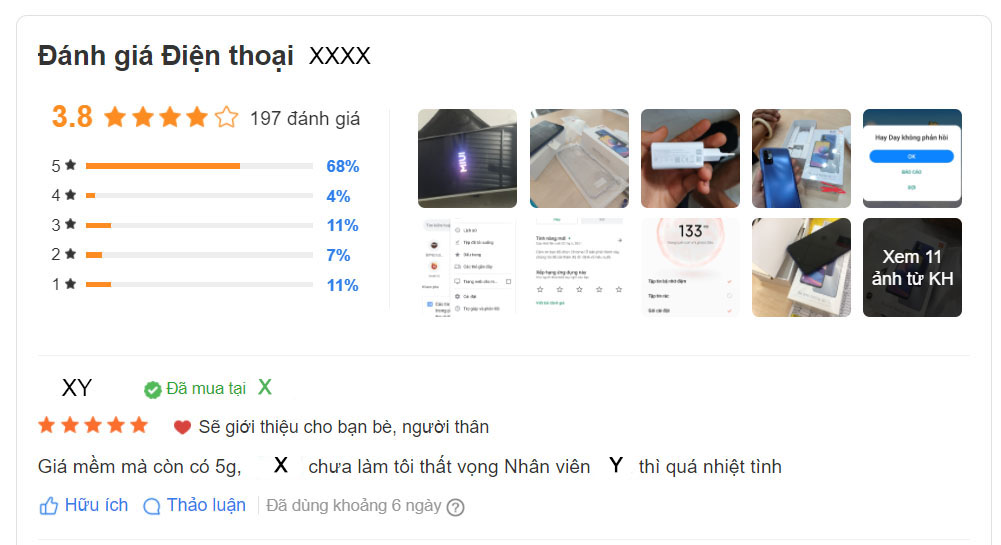}
    \caption {A feedback from an e-commerce site in Vietnam. The comment means "\textit{(I) will recommended to friends and family: Good prices but still have 5g, X has never made me disappointed. Y staff is very enthusiastic.}" in English.}
    \label{fig:1}
\end{figure}

User interface contributes a significant part to the shopping experience on e-commerce platforms. The user interfaces of e-commerce sites are more convenient than ever before with the help of ABSA techniques. If an e-commerce site adopts ABSA to their platform, customers can focus on corresponding reviews effectively by choosing the aspect-based sentiment text they care. Besides, the site owners can keep track of their product and service qualities with the help of ABSA. Several Chinese e-commerce platforms such as Taobao, Dianping deploy ABSA-based user interfaces to improve user experience. Therefore, the potential and importance of ABSA techniques for this area are immense. On the other hand, E-commerce sites present in Vietnam are still inferior in providing feedback to users. Most e-commerce platforms in Vietnam provide a simple feedback system: users leave their comments on the system along with a 5-star rating system like the one in Figure \ref{fig:1}. Such systematic platforms include thegioididong\footnote{\url{https://www.thegioididong.com/}}, fptshop\footnote{\url{https://fptshop.com.vn/}}, shopee\footnote{\url{https://shopee.vn/}}, tiki\footnote{\url{https://tiki.vn/}}, and lazada\footnote{\url{https://www.lazada.vn/}}. Different from the rest, foody\footnote{\url{https://www.foody.vn/}} (a restaurant review platform) allows users to respond on a 10-point scale and provides that score on several specific aspects (location, price, quality, service, and space). Therefore, we focus on the span detection for the ABSA problem, which not only detects aspects and their sentiment polarity but also detects spans of opinion.

\begin{figure*}[ht]
    \centering
    \includegraphics[scale=0.624]{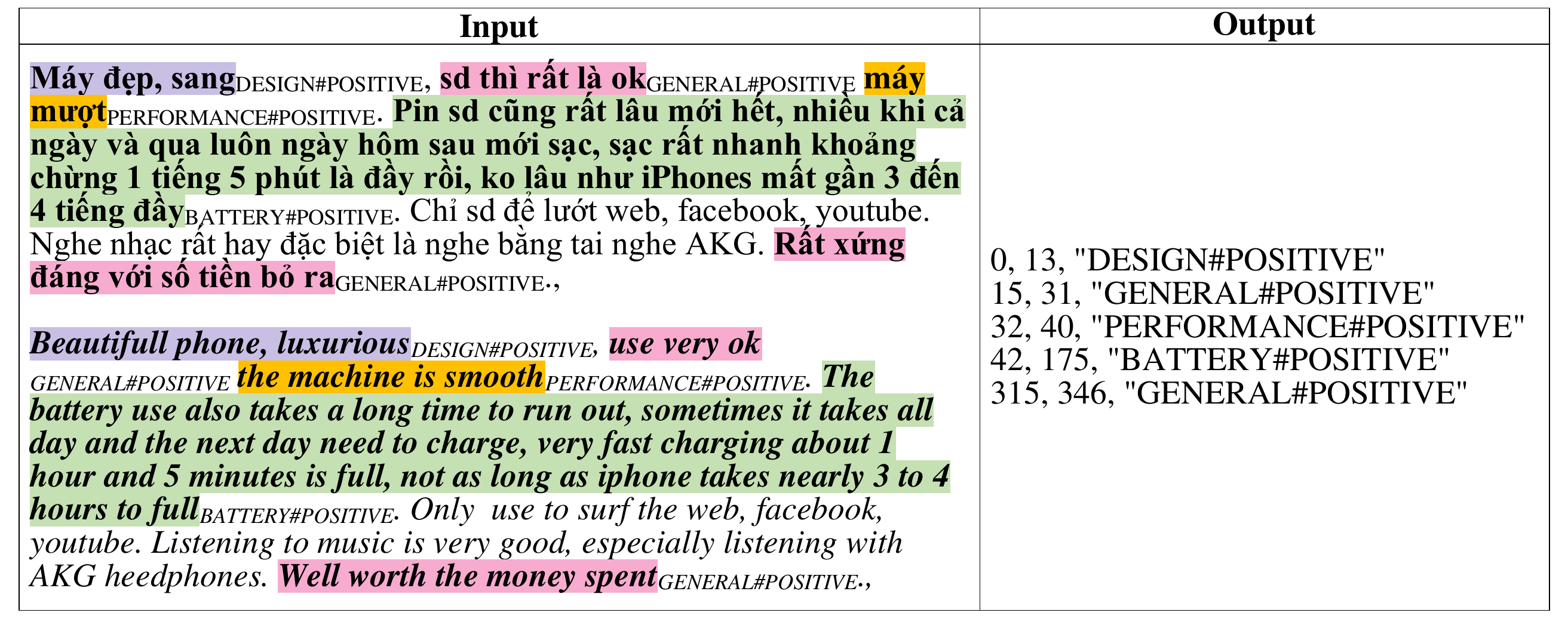}
    \caption {Examples illustrating spans for aspect-based sentiment analysis in Vietnamese.}
    \label{fig:2}
\end{figure*}

To the best of our knowledge, current public datasets are constructed for ABSA, which limits further explorations of span detection. To addressing the problem and advancing the related research, this paper presents UIT-ViSD4SA, a benchmark Vietnamese smartphone feedback dataset for ABSA and span detection. All the feedback in UIT-ViSD4SA is collected from an e-commerce platform. There are 11,122 user comments, and each is manually annotated according to its spans towards ten fine-grained aspect categories with sentiment polarities. Figure \ref{fig:2} shows an annotated illustrative datapoint.


We have three main contributions summarized as follows:  
\begin{itemize}
    \item First and foremost, we create a benchmark Vietnamese feedback dataset toward span detection aspect category sentiment analysis, named UIT-ViSD4SA, including 35,396 spans on 11,122 real-world smartphone feedback comments annotated with ten aspect categories. The dataset is available freely for research purposes.
    \item Next, we propose an approach using BiLSTM-CRF with embedding fusion for span detection in Vietnamese aspect-based sentiment analysis.
    \item Last but not least, we provide several case studies and future suggestions for span detection in Vietnamese aspect-based sentiment analysis.
\end{itemize}

The rest of the paper is organized as follows. In Section \ref{relatedwork}, we present the related work. In Section \ref{dataset}, we explain the data building process. The architecture of the approach is described in detail in Section \ref{approach}. In Section \ref{experiment}, we implement a BiLSTM-CRF model to solve the problem and analysis to find the weakness of the method on our dataset. Finally, Section \ref{conclusion} draws conclusions and future work.

\section{Related Work and Dataset}
\label{relatedwork}

The SemEval dataset series includes user reviews from e-commerce websites motivated for much-related ABSA research \cite{Li2019AUM,luo-etal-2020-grace,chen-qian-2020-relation}. The SemEval-2014 task 4 (SE-ABSA14) \cite{pontiki-etal-2014-semeval} dataset consists of restaurant and laptop reviews. The restaurant subset includes five aspects categories (i.e., Food, Service, Price, Ambience and Anecdotes/Miscellaneous) and four polarity labels (i.e., Positive, Negative, Conflict and Neutral). The laptop subset was just annotated for aspect category detection and sentiment polarity classification. SemEval-2015 Task 12 \cite{pontiki-etal-2015-semeval} dataset (SE-ABSA15) is built based on SE-ABSA14. SE-ABSA15 describes its aspect category as an entity type combined with an attribute type (e.g., Food\#Style Options) and removes the Conflict polarity label. The SemEval-2016 task-5 \cite{pontiki-etal-2016-semeval} dataset (SE-ABSA16)  extended SE-ABSA15 to new domains such as Hotels, Consumer Electronics, Telecom, Museums, and other languages (Dutch, French, Russian, Spanish, Turkish, and Arabic).

Compared with the prosperity of rich resource languages such as English, Chinese, or Spanish, the number of high-quality Vietnamese datasets are very low. In 2018, the first ABSA shared-task in Vietnamese was organized by the Vietnamese Language and Speech Processing (VLSP) community \cite{articlevlsp}. VLSP provided an ABSA dataset composed of hotel and restaurant reviews. Unfortunately, the VLSP dataset inspired by SE-ABSA15 was only annotated for entity\#atribute aspect category and its sentiment but ignoring the Opinion Target Extraction. Nguyen et al. \shortcite{8919448} proposed the dataset on the same domains as restaurants and hotels, including only 7,828 reviews at document-level with seven aspects combined with five polarity sentiments for two tasks. Dang et al. \shortcite{10.1145/3446678} also built a dataset for the same domain as two previous works annotated with high inter-annotator agreements at the sentence level. Mai et al. \shortcite{inbook} collected and annotated Vietnamese ABSA corpora consisted of only 2,098 sentences for two tasks: opinion target extraction and sentiment polarity detection for the smartphone domain. They presented a multi-task model for the two tasks using the sequence labeling scheme associated with bidirectional recurrent neural networks (BRNN) and conditional random field (CRF). To evaluate aspect-based sentiment analysis for mobile e-commerce, Phan et al., \shortcite{phan2021sa2sl} created a benchmark dataset (UIT-ViSFD) with 11,122 comments based on a strict annotation scheme. Furthermore, they developed a social listening system in Vietnamese based on aspect-based sentiment analysis.
\begin{table*}[ht]
\centering
\resizebox{\textwidth}{!}{
\begin{tabular}{|l|l|}
\hline
\multicolumn{1}{|c|}{\textbf{Aspect}} & \multicolumn{1}{c|}{\textbf{Definition}}                                                                  \\ \hline
\textbf{SCREEN}                       & User comments   express screen quality, size, colors, and display technology.                              \\ 
\textbf{CAMERA}                       & The comments   mention the quality of a camera, vibration, delay, focus, and image colors.                \\ 
\textbf{FEATURES}                     & The users refer to features, fingerprint sensor, wifi connection, touch and face detection of   the phone. \\ 
\textbf{BATTERY}                      & The comments   describes battery capacity and battery quality.                                             \\ 
\textbf{PERFORMANCE}                  & The reviews   describe ramming capacity, processor chip, performance using, and smoothness of the phone.  \\ 
\textbf{STORAGE}                      & The comments mention   storage capacity, the ability to expand capacity through memory cards.              \\ 
\textbf{DESIGN}                       & The reviews refer   to the style, design, and shell.                                                       \\ 
\textbf{PRICE}                        & The comments   present the specific price of the phone.                                                   \\ 
\textbf{GENERAL}                      & The reviews of   customers generally comment about the phone.                                             \\ 
\textbf{SER\&ACC$^7$}                     & The comments   mention sales service, warranty, and review of accessories of the phone.                   \\ \hline
\multicolumn{2}{l}{\footnotesize $^7$ SER\&ACC is short for SERVICE and ACCESSORIES.}    
\end{tabular}}
\caption{The full list of ten aspects and their definitions.}
\label{tab:1}
\end{table*}
For span detection in ABSA, Hu et al. \shortcite{hu-etal-2019-open} proposed a span-based extract-then-classify framework, where multiple opinion targets are directly extracted from the sentence under the supervision of target span boundaries, and corresponding polarities are then classified using their span representations. This work is inspired by advances in machine comprehension and question answering \cite{seo2018bidirectional,xu-etal-2018-double}, where the task is to extract a continuous span of text from the document as the answer to the question \cite{rajpurkar2016squad,nguyen2020vietnamese}. 
Xu et al. \shortcite{xu-etal-2020-aspect} presented a neat and effective multiple CRFs based structured attention model capable of extracting aspect-specific opinion spans. The sentiment polarity of the target is then classified based on the extracted opinion features and contextual information.


\section{Dataset Creation and Analysis}
\label{dataset}
Based on the benchmark dataset proposed by Phan et al. \shortcite{phan2021sa2sl}, we develop a new dataset for span detection for ABSA in Vietnamese. The creation process of our dataset is described as follows. To begin with, we edit and revise the annotation guidelines from \cite{phan2021sa2sl}  for annotators to determine spans and how to annotate data correctly (see Section \ref{spandefinition}). Annotators are trained with the guidelines and annotate data to ensure that the F1-score in the training process reaches over 80\% before performing data annotation independently (see Section \ref{annotationprocess}). Finally, we provide an analysis of the dataset that helps experts understand this dataset (see Section \ref{dataanalysis}).
 
We utilize the ABSA dataset collected from an e-commerce website for smartphones in Vietnam, which allows customers to write fine-grained reviews of a smartphone they have purchased. In the reviews, users comment on multiple aspects either explicitly or implicitly about many aspects such as camera, price, battery, service, and etcetera. The dataset includes 11,122 feedback with four attributes: comment, n\_star, date\_time, and label. Table \ref{tab:1} summarizes ten aspects in the guidelines, and each aspect has one of three sentiments (positive, negative, and neutral). 


\subsection{Span Definition and Annotation Guidelines}
\label{spandefinition}

Following the annotation guidelines proposed by Phan et al. \shortcite{phan2021sa2sl}, we add some definitions and rules to form the core of data construction. We reuse the ten predefined aspect categories as in Table \ref{tab:1}, with each aspect category mentioned within the review, the sentiment polarity over the aspect category is labeled as Positive, Neutral, or Negative. The span is defined as the shortest span containing the opinions of the user about the aspect category. With ten predefined aspects, annotators are asked to annotate spans towards aspect categories with sentiment polarities of each review. Suppose a review is given, when a span is discovered within the review either explicitly or implicitly, the aspect category with sentiment polarity of that span is labeled as aspect\#polarity as in Figure \ref{fig:1}.

\subsection{Annotation Process}
\label{annotationprocess}

\begin{figure}[H]
    \centering
    \includegraphics[scale=0.42]{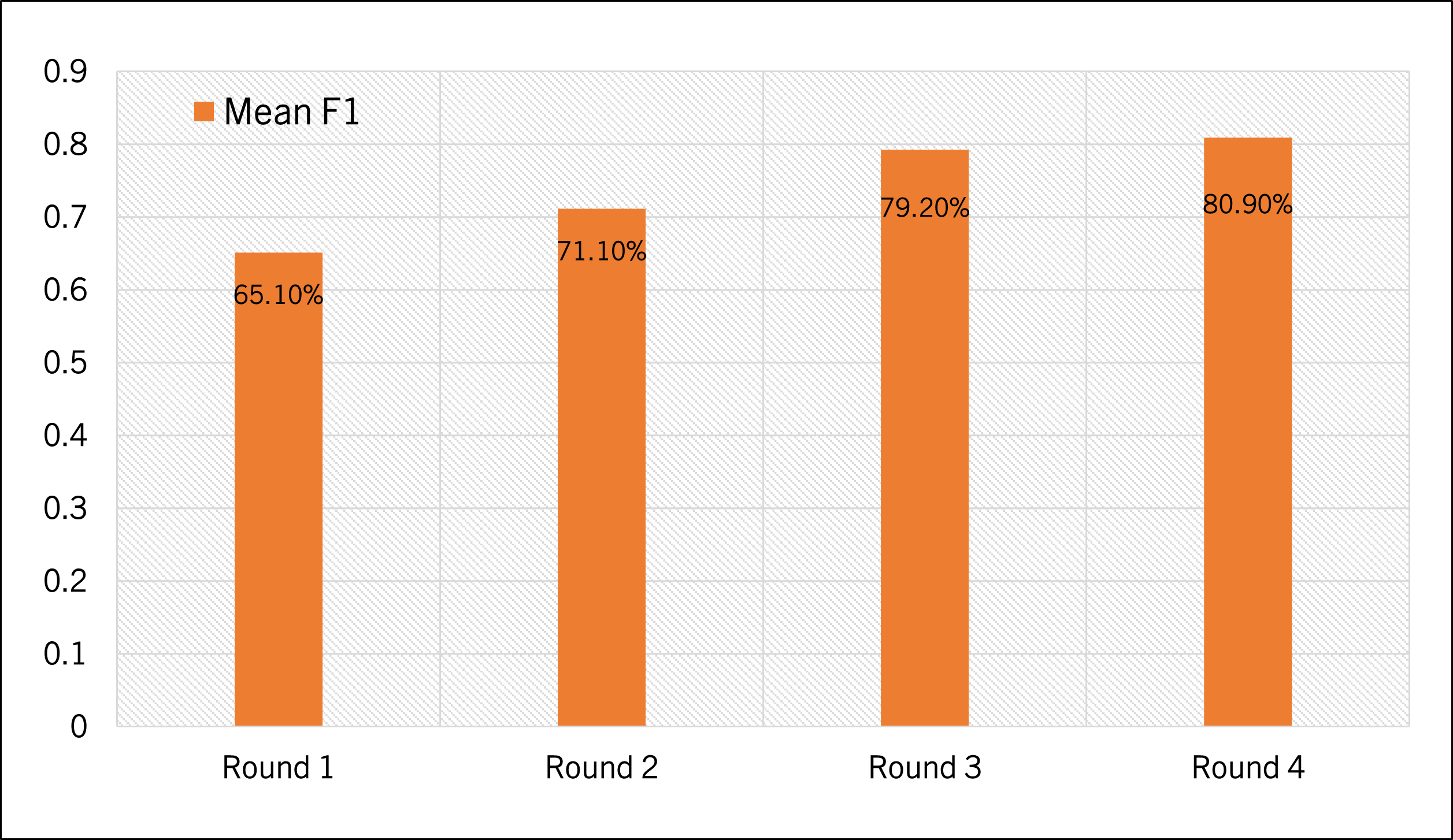}
    \caption {Results for four rounds of measurement of F1-score.}
    \label{fig:x}
\end{figure}

Three phases of annotation are conducted as follows. To begin with, we train annotators with the guidelines and randomly take about 30-70 reviews in the dataset to annotate, then calculate F1-core per review for those annotated data. For disagreement cases, annotators decide the final label by discussing and having a voting poll. Annotators spend four training rounds to obtain a high F1-score above 80\% before performing data annotation independently. Figure \ref{fig:x} shows the F1-score during training phases.
An annotation is a triple $(d, l, o)$, where $d$ is a document id, $l$ a label, and $o$ is a list of start-end character offset tuples. An annotator $i$ contributes a (multi)set $A_{i}$ of (token) annotations. We compute (1) for each 2-combination of annotators and report arithmetic mean of F1 across all these combinations \cite{Hripcsak2005TechnicalBA}. Grouping annotations by documents or labels allows us to calculate F1 per document or label. 
\begin{align} 
F1_{ij}=\frac{2\times|A_{i}\cap A_{j}|}{A_{i}+A_{j}} 
\end{align}
Finally, our dataset is divided randomly into three sets: the training (Train), development (Dev), and test (Test) in the ratio 7:1:2. Figure \ref{fig:1} presents an example review of our dataset and corresponding annotations.
\begin{figure*}[ht]
    \centering
    \includegraphics[scale=0.3]{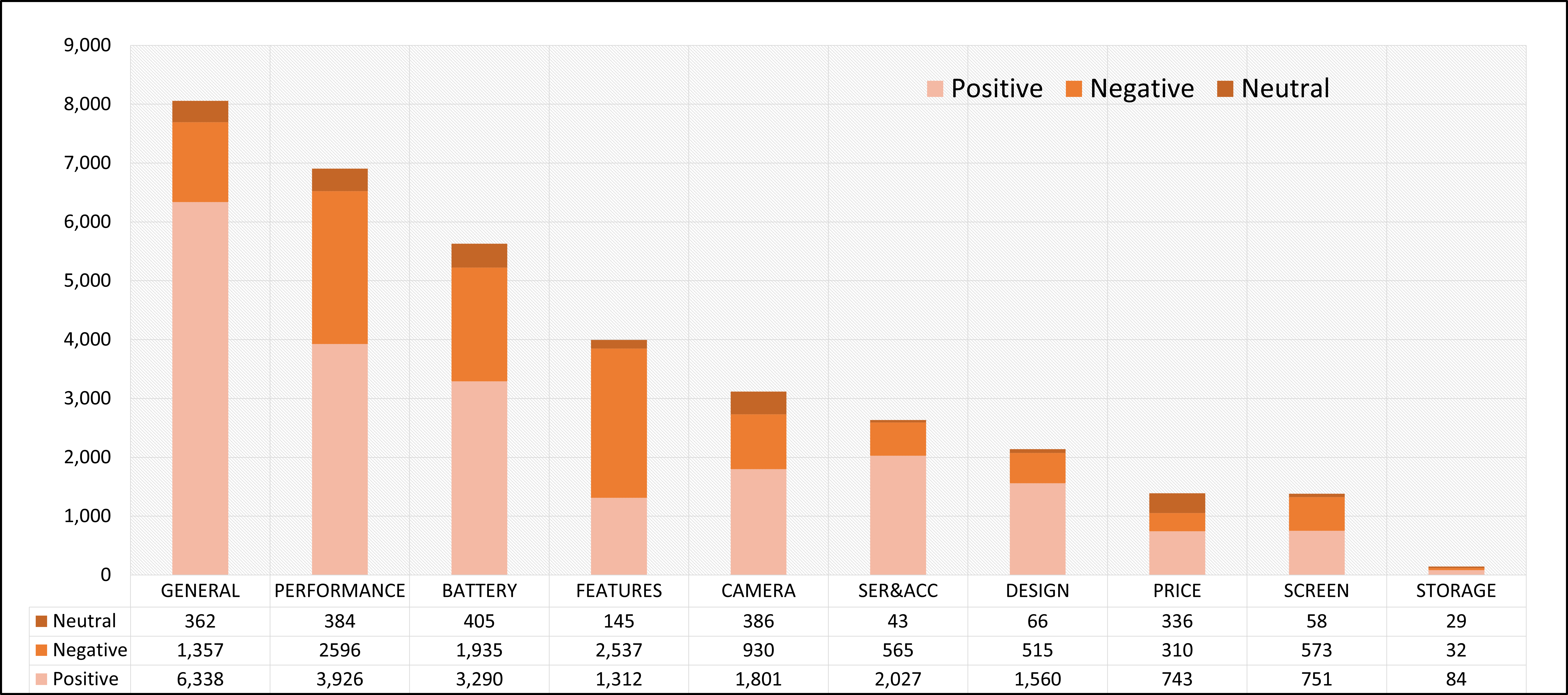}
    \caption {The distribution of 10 fine-grained aspect categories.}
    \label{fig:3}
\end{figure*}
\begin{table*}[ht]
\centering
\begin{tabular}{|l|r|r|r|r|r|r|r|}
\hline
\multicolumn{1}{|c|}{\bf Set} & \multicolumn{1}{c|}{\bf Comment} & \multicolumn{1}{c|}{\bf \begin{tabular}[c]{@{}c@{}}Average aspect \\ per comment\end{tabular}} & \multicolumn{1}{c|}{\bf \begin{tabular}[c]{@{}c@{}}Average \\ span length\end{tabular}} & \multicolumn{1}{c|}{\bf Positive} & \multicolumn{1}{c|}{\bf Negative} & \multicolumn{1}{c|}{\bf Neutral} & \multicolumn{1}{c|}{\bf Total span} \\ \hline
Train                     & 7,784                        & 3.2                                                                                       & 32.6                                                                                & 15,356                        & 7,793                         & 1,560                        & \multirow{3}{*}{35,396}         \\ 
Dev                       & 1,113                        & 3.1                                                                                       & 32.4                                                                                & 2,110                         & 1,144                         & 241                          &                                 \\ 
Test                      & 2,225                        & 3.2                                                                                       & 32.5                                                                                & 4,266                         & 2,269                         & 413                          &                                 \\ \hline
\end{tabular}
\caption{The overview statistics of our UIT-ViSD4SA dataset.}
\label{tab:3}
\end{table*}

\subsection{Dataset Analysis}
\label{dataanalysis}
Figure \ref{fig:3} presents the distribution of ten aspect categories in our dataset UIT-ViSD4SA. People tend to give a smartphone an overall rating, with 22.76\% of reviews mentioning GENERAL. Users frequently pay great attention to aspects related to their needs, such as PERFORMANCE, BATTERY, FEATURES, and CAMERA. 

The statistics of our dataset are presented in Table \ref{tab:3}. Our dataset includes 35,396 spans over 11,122 comments. Through our analysis, the dataset has an uneven distribution of sentiment labels. The positive polarity accounts for the most significant number of labels, followed by the negative polarity and neutral polarity. On average, the reviews have three spans, with each span being about 32 characters long. We hope our dataset will open the new shared task for evaluating span detection in aspect-based sentiment analysis.

\section{Our Approach}
\label{approach}

\begin{figure}[H]
    \centering
    \includegraphics[scale=0.42]{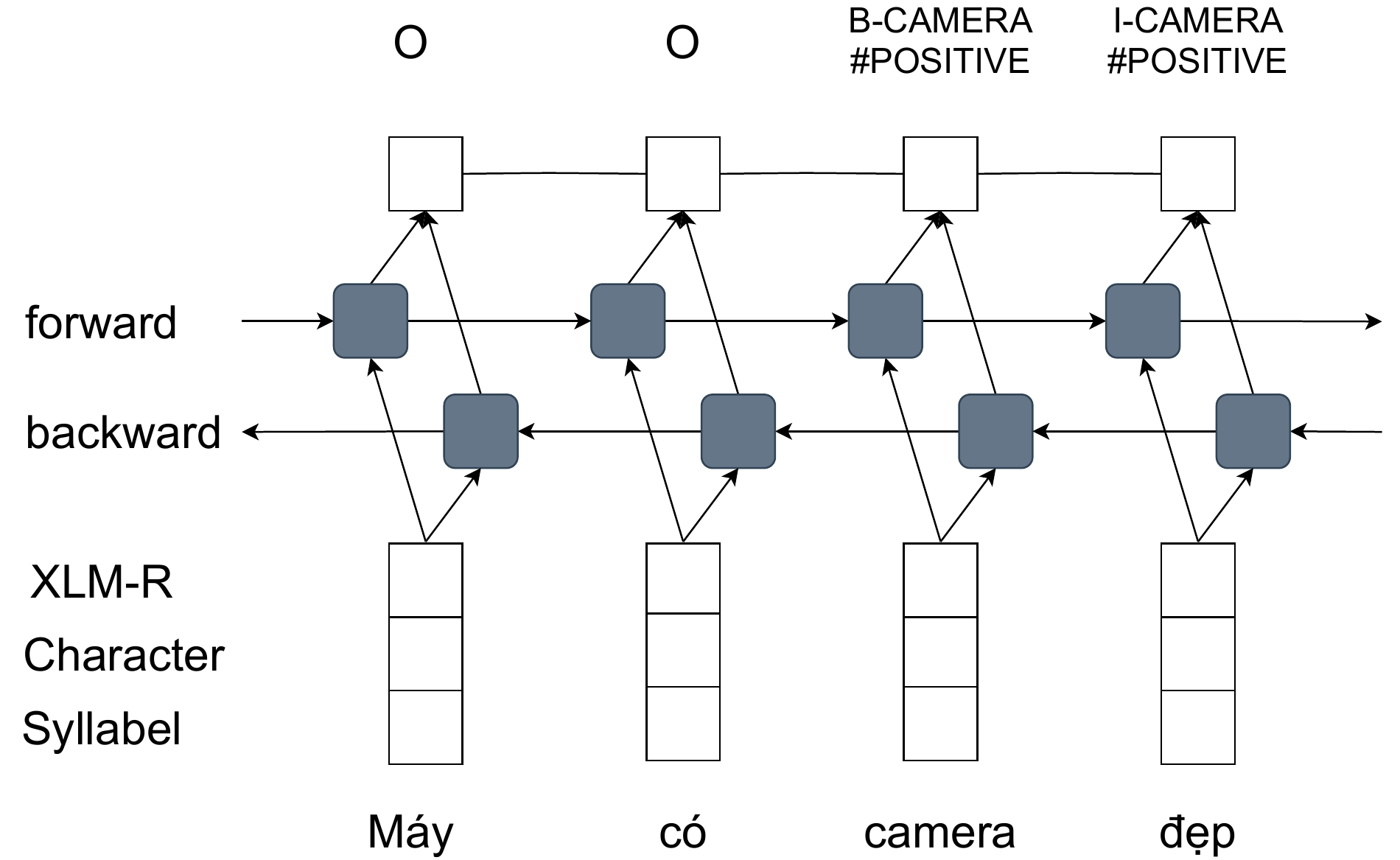}
    \caption {BiLSTM-CRF network with embedding layers (the example feedback means \textit{"This phone has a good camera"} in English).}
    \label{fig:4}
\end{figure}

For the baseline evaluation, we consider span detection for ABSA as a sequence labeling problem at the syllable level.
We employ a BiLSTM-CRF model \cite{huang2015bidirectional} with embedding fusion to solve the task. The BiLSTM-CRF model comprises three layers: token embedding layer giving contextualized vector representation of input sequence, passed into the BiLSTM-CRF sequence labeler as depicted in Figure \ref{fig:4}.


\subsection{Embedding Fusion Layer}
The embedding layer takes as input a sequence of $N$ tokens $(x_{1}, x_{2},.., x_{N} )$, and output a fixed-dimensional vector representation of each token $(e_{1}, e_{2}, ..., e_{N} )$. We use an embedding fusion of syllable embedding \cite{nguyen-etal-2017-word}, character embedding (CharLSTM), contextual embedding from XLM-RoBERTa \cite{conneau2020unsupervised}.



\subsection{Bidirectional Long Short-Term Memory (BiLSTM)}
A long-short term memory network (LSTM) is a special type of Recurrent neural network (RNN) introduced by Hochreiter et al.,\shortcite{Hochreiter1997LongSM}, which can capture a long-distance semantic relationship by maintaining a memory cell store context information. LSTMs do not suffer from vanishing and exploding gradient problems. The LSTM is equipped with a memory cell with an adaptive adjustment mechanism that adjusts information to be added to or removed from the cell. The memory cell is continuously updated during encryption, and the information rate is determined by three kernel gates, including input, forget and output. In terms of formality, the encryption process at the time step \textit{t} is performed as follows:
 
\begin{align}
\centering
\begin{split}
    i_{t} = \sigma(W_{hi}h_{t-1}+W_{ei} e_{t}^{w}+b_{i})
\end{split}\\
\begin{split}
    f_{t} = \sigma(W_{hf}h_{t-1}+W_{ef} e_{t}^{w}+b_{f})
\end{split}\\
\begin{split}
    \widetilde{c_{t}} = tanh(W_{hc}h_{t-1}+W_{ec} e_{t}^{w}+b_{c})
\end{split}\\
\begin{split}
    c_{t} = f_{t}\odot c_{t-1}+i_{t}\odot\widetilde{c_{t}}
\end{split}\\
\begin{split}
    o_{t} = \sigma(W_{ho}h_{t-1}+W_{eo} e_{t}^{w}+b_{o}
\end{split}\\
\begin{split}
    h_{t} = o_{t}\odot tanh(c_{t})
\end{split}
\end{align}

where $c_{t}$, $i_{t}$, $f_{t}$, and $o_{t}$ represent the memory cell, input gate, forget gate and output gate respectively. $e_{t}^{w}$ and $h_{t}$ donate the word embedding vector and hidden state vector at time $t$. Both $\sigma$ and $tanh$ are the activation functions, and $\odot$ represents the element-wise product. $W\ast$ and $b\ast$ are network parameters that donate the weight matrices and bias vectors. Although LSTM can solve the long-distance dependency problem, it still loses some semantic information due to the sequential encoding way of LSTM. For example, $h_{t}$ only contains the semantic information before time step $t$. Therefore, a Bidirectional LSTM (BiLSTM) is needed to model both the forward and backward context information as in equation (8,9), and the two hidden states are concatenated to obtain the final output as equation (10):
\begin{align}
\begin{split}
    \overrightarrow{h_{t}}=F(e^{w}_{t},\overrightarrow{h_{t-1}})
\end{split}\\
\begin{split}
    \overleftarrow{h_{t}}=F(e^{w}_{t},\overleftarrow{h_{t-1}})
\end{split} \\
\begin{split}
    h_{t}=[\overrightarrow{h_{t}},\overleftarrow{h_{t}}]
\end{split}
\end{align}

\subsection{Conditional Random Fields (CRF)}

Conditional Random Fields (CRF) \cite{10.5555/645530.655813} is a sequence modeling framework that brings in all the advantages of MEMMs models \cite{McCallum2000,ratnaparkhi-1996-maximum} while also solving the label bias problem. With CRF, the inputs and outputs are directly connected, unlike LSTM and BiLSTM networks where memory cells/recurrent components are employed. Given a training dataset $D=(x^{1},y^{1}),...,(x^{N},y^{N})$ of $N$ data sequences to be labeled $x^{i}$ and their corresponding label sequences $y^{i}$, CRF maximizes the conditional log-likelihood of label sequences based on the data sequences as shown as follow: 

\begin{align}
    L=\sum_{i=1}^{N}log(P(y^{i}|x^{i}))-\sum_{k=1}^{K}\frac{\lambda _{k}^{2}}{2\sigma^{2}}
\end{align}

\section{Experiments and Results}
\label{experiment}
\begin{table*}[ht]
\resizebox{\textwidth}{!}{
\begin{tabular}{|l|r|r|r|r|r|r|}
\hline
{\bf System}                                            & \multicolumn{1}{c|}{\bf P$_{Micro}$} & \multicolumn{1}{c|}{\bf R$_{Micro}$} & \multicolumn{1}{c|}{\bf F1$_{Micro}$} & \multicolumn{1}{c|}{\bf P$_{Macro}$} & \multicolumn{1}{c|}{\bf R$_{Macro}$} & \multicolumn{1}{c|}{\bf F1$_{Macro}$} \\ \hline
Aspect (syllable)                            & 64.55                         & 60.86                         & 62.65                         & 62.76                         & 57.28                         & 59.74                         \\ 
Aspect (syllable + char)                            & 63.78                         & 62.11                         & 62.93                         & 61.64                         & 58.91                         & 60.21                         \\ 
Aspect (syllable + char + XLM-R$_{Base}$)            & 65.63                         & 65.15                         & 65.39                         & 62.88                         & 61.62                         & 62.17                         \\ 
Aspect (syllable + char + XLM-R$_{Large}$)           & 64.96                         & 66.85                         & \textbf{65.89}                         & 62.00                         & 63.56                         & \textbf{62.76}                         \\ \hline
Polarity (syllable)                          & 52.36                         & 50.10                         & 51.20                         & 46.71                         & 38.37                         & 41.05                         \\ 
Polarity (syllable + char)                    & 52.12                         & 51.00                         & 51.55                         & 44.44                         & 38.79                         & 40.68                         \\ 
Polarity (syllable + char + XLM-R$_{Base}$)           & 54.88                         & 55.91                         & 55.39                         & 46.87                         & 46.39                         & 46.57                         \\ 
Polarity (syllable + char  + XLM-R$_{Large}$)         & 56.89                         & 59.78                         & \textbf{58.30}                         & 49.00                         & 50.60                         & \textbf{49.77}                         \\ \hline
Aspect-polarity (syllable)                  & 61.87                         & 54.55                         & 57.98                         & 48.77                         & 34.27                         & 37.64                         \\
Aspect-polarity (syllable + char)               & 59.51                         & 57.56                         & 58.52                         & 43.66                         & 37.53                         & 39.30                         \\ 
Aspect-polarity (syllable + char + XLM-R$_{Base}$)  & 60.71                         & 61.62                         & 61.16                         & 46.18                         & 43.42                         & 44.37                         \\ 
Aspect-polarity (syllable + char + XLM-R$_{Large}$)  & 61.78                         & 62.99                         & \textbf{62.38}                         & 46.84                         & 45.46                         & \textbf{45.70 }                        \\ \hline
\end{tabular}}
\caption{The overall experimental results.}
\label{tab:4}
\end{table*}
\begin{table}[ht]
\centering
\resizebox{\columnwidth}{!}{
\begin{tabular}{|l|r|r|r|}
\hline
{\bf Aspect}            & \multicolumn{1}{c|}{\bf Precision} & \multicolumn{1}{c|}{\bf Recall} & \multicolumn{1}{c|}{\bf F1-score} \\ \hline
BATTERY     & 71.04                          & 73.58                       & 72.29                         \\ 
CAMERA      & 75.09                          & 77.82                       & \textbf{76.43}                            \\ 
DESIGN      & 68.13                          & 70.66                       & 69.37                         \\ 
FEATURES    & 58.76                          & 59.34                       & 59.05                          \\ 
GENERAL     & 64.74                          & 68.90                       & 66.76                         \\ 
PERFORMANCE & 62.37                          & 63.11                       & 62.74                         \\ 
PRICE       & 46.72                          & 47.98                       & 47.35                         \\ 
SCREEN      & 65.83                          & 68.70                       & 67.23                         \\ 
SER\&ACC    & 65.18                          & 61.83                       & 63.46                         \\ 
STORAGE     & 45.16                             & 46.67                       & 45.90                         \\ \hline
\end{tabular}}
\caption{Result per class for only aspect label.}
\label{tab:5}
\end{table}
\begin{table}[ht]
\centering
\begin{tabular}{|l|r|r|r|}
\hline
{\bf Sentiment}         & \multicolumn{1}{l|}{\bf Precision} & \multicolumn{1}{l|}{\bf Recall} & \multicolumn{1}{l|}{\bf F1-score} \\ \hline
NEGATIVE & 47.05                          & 47.56                       & 47.30                         \\ 
NEUTRAL  & 36.57                          & 35.97                       & 36.26                         \\ 
POSITIVE & 63.52                          & 68.50                       & \textbf{65.92 }                        \\ \hline
\end{tabular}
\caption{Result per class for only sentiment polarity label.}
\label{tab:6}
\end{table}
\begin{table}[ht]
\centering
\resizebox{\columnwidth}{!}{
\begin{tabular}{|l|r|r|r|}
\hline
{\bf Aspect}            & \multicolumn{1}{c|}{\bf Negative} & \multicolumn{1}{c|}{\bf Neutral} & \multicolumn{1}{c|}{\bf Positive} \\ \hline
BATTERY     & 54.62                         & 44.07                        & \textbf{78.40}                      \\ 
CAMERA      & 58.97                         & 55.65                        & 77.54                         \\ 
DESIGN      & 46.15                         & 00.00                        & 75.75                         \\ 
FEATURES    & 50.73                         & 22.22                         & 68.11                         \\ 
GENERAL     & 52.12                         & 52.73                        & 67.87                         \\ 
PERFORMANCE & 45.87                         & 24.19                        & 70.84                         \\ 
PRICE       & 32.69                         & 15.05                        & 52.63                         \\ 
SCREEN      & 48.62                         & 46.15                          & 71.13                         \\ 
SER\&ACC    & 22.56                         & 00.00                        & 72.17                         \\ 
STORAGE     & 15.38                         & 00.00                          & 57.14                         \\ \hline
\end{tabular}}
\caption{F1-score per class for aspect\#polarity label.}
\label{tab:7}
\end{table}
\subsection{Experimental Settings}

Following the IOB format (short for inside, outside, beginning), our dataset is converted with data containing only aspect labels (SCREEN, BATTERY, CAMERA, etcetera.), sentiment labels only (POSITIVE, NEUTRAL, and NEGATIVE), and data containing both aspect and sentiment labels (SCREEN\#POSITIVE, BATTERY\#NEGATIVE, etcetera.) to evaluate our approach comprehensively. Our word embeddings have three parts: syllable (1), character (2), and contextual from XLM-R(3), with an embedding dimension of 100. We set the hidden layers of LSTM as 400, the dropout rate as 0.33, and the batch size as 5,000 with 30 epochs for training. All experiments are conducted on a single NVIDIA T4 GPU card.
\subsection{Evaluation Metrics}
In this paper, we use three evaluation metrics: Precision, Recall, and F1-score. A predicted span is correct only if it exactly matches the gold standard span. To gain a comprehensive view, we calculate these evaluation metrics on both the micro and macro averages.


\subsection{Experimental Results}
Table \ref{tab:4} presents performances of the BiLSTM-CRF model with three types of embedding fusion on the aspect, polarity, aspect\#polarity span detection. According to our results, we can see that concatenate three embedding layers (syllable, character, and bert-based embedding) have significantly better performance than just one or two embedding layers. In particular, syllabel+char+XLMRlarge achieves the best F1$_{macro}$ of 62.76\%, 49.77\%, and 45.70\% for aspect, polarity, and aspect\#polarity, respectively, whereas the model with just syllable embedding layer shows the lowest performances. On the other hand, our method tends to be less efficient with labels which consist of polarity, in which polarity task reach 49.77\% F1$_{macro}$ while aspect\#polarity task gets  45.70\% F1$_{macro}$.

Detailed results per class of each task are shown in Tables \ref{tab:5}, \ref{tab:6}, and \ref{tab:7} (with aspect\#polarity label, we only show F1-score). For aspect task, only two aspects have a high F1-score above 70\% (CAMERA and BATTERY) while the rest range from 60-70\%, especially F1-score of PRICE and STORAGE is relatively low (below 50\%). With the polarity task, the result is descending with the order POSITIVE, NEGATIVE, NEUTRAL. The result of aspect\#polarity can be considered the sum of the two previous tasks: previous high-performing aspects labels combined with positive give the highest result. This result explains the lack of quantity uniformity in the labels (labels consist of NEUTRAL polarity only cover 6.25\% of our dataset, detail in Figure \ref{fig:3}). In general, our approach gets better performance when it comes to detecting span for aspect than polarity and aspect\#polarity span detection. However, their ability to detect span for all types of labels is still limited (F1-score below 80\%), which will be exploited in future work.
\begin{figure*}[ht]
    \centering
    \includegraphics[scale=0.59]{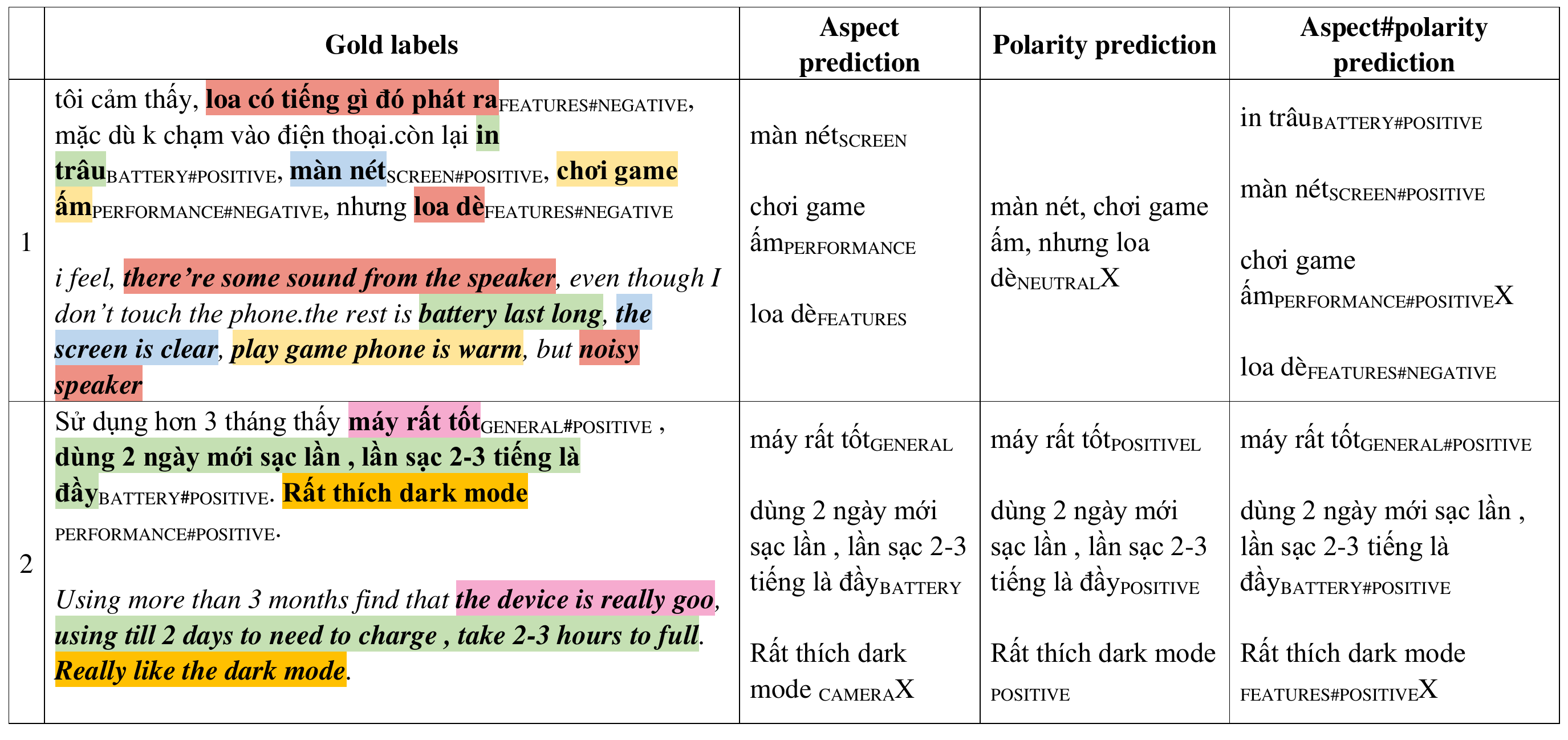}
    \caption {Case study. The spans are bold with aspects and their polarities are given as subscripts. Incorrect predictions are marked with X.}
    \label{fig:5}
\end{figure*}


\subsection{Case Study}
\label{casestudy}

Figure \ref{fig:5} shows several cases predicted by the BiLSTM-CRF model. After reviewing the cases, we found that the model commits three common types of errors that can not detect spans, misclassify the sentiment polarity, and detect the wrong boundary of spans. As observed in the first sentence, both three types of models can not detect the span \textit{"there's some sound from the speaker"}. With the cases of misclassification, we found that many cases of this mistake contained English loanwords. For example, in comment 2, the span \textit{"Really like the dark mode"} is about the interface, and we annotate it as PERFORMACNE\#POSITIVE. However, the model can understand it and classify it as CAMERA (aspect label model) or FEAUTURE\#POSITIVE (aspect\#polarity label model). This feature needs attention and research in future studies because the Vietnamese language feature (especially in technology) often includes many loanwords with meanings that can be similar or different from the original language. Besides, the polarity model incorrectly predicts the target span by detecting the whole span \textit{"the screen is clear, play game phone is warm, but noise speaker"} as a NEUTRAL span. This mistake can be blamed on the way we train the model just with polarity label, which makes it difficult for the model to identify the aspect to which the emotional label is directed. The proof for this argument is that the model with the label aspect\#polarity can detect the boundary of spans better than the polarity model.

\section{Conclusion and Future Work}
\label{conclusion}

This paper presented UIT-ViSD4SA, which is a new dataset for span detection on aspect-based sentiment analysis and consists of over 35,000 human-annotated spans on 11,122 comments for mobile e-commerce. Each feedback is manually annotated according to its spans towards ten fine-grained aspect categories with their sentiment polarities. BiLSTM-CRF uses an embedding fusion  of syllable, character, and contextual embedding, which had the highest 62.76\% F1$_{macro}$ for span detection on aspect, 49.77\% F1$_{macro}$ for span detection on polarity, and 45.70\% F1$_{macro}$ for span detection on aspect\#polarity, respectively. In general, the performances are relatively not high and challenging for further machine learning-based models. We hope the release of UIT-ViSD4SA could motivate the development of machine learning models and applications.

In future work, we give several directions: (1) Inspired by Yuan et al. \shortcite{yuan2020enhancing}, multilingual pre-trained language models can be used for enhancing span boundary detection. (2) Improving the performance of this task can be used with approaches based on machine comprehension reading, and other approaches \cite{hu-etal-2019-open,xu-etal-2020-aspect}. (3) Inspired by Xu et al. \shortcite{xu2019bert}, review reading comprehension for Vietnamese can be developed on our dataset. (4) Span detection is a challenging task that can motivate various future works on constructive analysis \cite{fujita2019dataset,10.1007/978-3-030-79457-6_49}, emotion analysis \cite{sosea2020canceremo,ho2019emotion}, complaint analysis \cite{preotiuc-pietro-etal-2019-automatically,nguyen2021vietnamese}, and opinion mining \cite{nguyen2018vlsp,jiang2019challenge}.


\bibliographystyle{acl}
\bibliography{references}

\end{document}